\documentclass[letterpaper, 10 pt, journal, twoside]{IEEEtran}

\IEEEoverridecommandlockouts                              

\usepackage{amsmath} 
\usepackage{amssymb}  
\usepackage{cite}
\usepackage{booktabs}
\usepackage[subnum]{cases}
\usepackage{mathtools}
\usepackage{balance}
\usepackage{color}
\usepackage{tabularx}
\usepackage{graphicx,dblfloatfix} 
\usepackage{listings}
\usepackage{textcomp}
\usepackage{url}
\usepackage{multirow}
\usepackage{algorithm}
\usepackage[noend]{algpseudocode}
\usepackage{ragged2e} 
\newcolumntype{Y}{>{\RaggedRight\arraybackslash}X} 

\usepackage{bm}
\usepackage[colorinlistoftodos]{todonotes}
\usepackage{hyperref}
\usepackage[normalem]{ulem}
\usepackage{bbm}
\usepackage{soul}

\usepackage[printonlyused]{acronym}

\acrodef{RSL}{Robotic Systems Lab}
\acrodef{COM}{center of mass}
\acrodef{SQP}{sequential quadratic problem}
\acrodef{VMC}{virtual model controller}
\acrodef{w.r.t.}{with respect to}
\acrodef{DOF}{degrees of freedom}
\acrodef{ZMP}{zero moment point}
\acrodef{IMU}{inertial measurement unit}
\acrodef{COT}{cost of transport}
\acrodef{JPL}{Jet Propulsion Laboratory}
\acrodef{HAA}{hip adduction/abduction}
\acrodef{HFE}{hip flexion/extension}
\acrodef{KFE}{knee flexion/extension}
\acrodef{ZMP}{zero-moment point}
\acrodef{QP}{quadratic programming}
\acrodef{SQP}{sequential quadratic programming}
\acrodef{WBC}{whole-body controller}
\acrodef{HO}{hierarchical optimization}
\acrodef{NLP}{nonlinear programming}
\acrodef{MPC}{model predictive control}

\newcommand{\deleted}[1]{}

\title{
Keep Rollin' -- Whole-Body Motion Control and Planning for Wheeled Quadrupedal Robots
}

\author{Marko Bjelonic, C. Dario Bellicoso, Yvain de Viragh, Dhionis Sako, \\ F. Dante Tresoldi, Fabian Jenelten and Marco Hutter
\thanks{Manuscript received: September, 10, 2018; Revised: December, 11, 2018; Accepted: January, 23, 2019.}
\thanks{This paper was recommended for publication by Editor Nikos Tsagarakis upon evaluation of the Associate Editor and Reviewers' comments. This work has been conducted as part of ANYmal Research, a community to advance legged robotics. This work was supported in part by the Swiss National Science Foundation through the National Centres of Competence in Research Robotics (NCCR Robotics) and Digital Fabrication (NCCR dfab).}%
\thanks{Correspondence should be addressed to Marko Bjelonic.}%
\thanks{All authors are with the Robotic Systems Lab, ETH Z\"urich, 8092 Z\"urich, Switzerland, email: \texttt{firstname.surname@mavt.ethz.ch}}
\thanks{Digital Object Identifier (DOI): see top of this page.}
}

\begin{document}

\markboth{IEEE Robotics and Automation Letters. Preprint Version. Accepted January, 2019}{Bjelonic \MakeLowercase{\textit{et al.}}: Keep Rollin' -- Whole-Body Motion Control and Planning for Wheeled Quadrupedal Robots}

\maketitle

\begin{abstract}
We show dynamic locomotion strategies for wheeled quadrupedal robots, which combine the advantages of both walking and driving. The developed optimization framework tightly integrates the additional degrees of freedom introduced by the wheels. Our approach relies on a zero-moment point based motion optimization which continuously updates reference trajectories. The reference motions are tracked by a hierarchical whole-body controller which computes optimal generalized accelerations and contact forces by solving a sequence of prioritized tasks including the nonholonomic rolling constraints. Our approach has been tested on ANYmal, a quadrupedal robot that is fully torque-controlled including the non-steerable wheels attached to its legs. We conducted experiments on flat and inclined terrains as well as over steps, whereby we show that integrating the wheels into the motion control and planning framework results in intuitive motion trajectories, which enable more robust and dynamic locomotion compared to other wheeled-legged robots. Moreover, with a speed of 4\,m/s and a reduction of the cost of transport by 83\,\% we prove the superiority of wheeled-legged robots compared to their legged counterparts.
\end{abstract}

\begin{IEEEkeywords}
Legged Robots, Wheeled Robots, Motion Control, Motion and Path Planning, Optimization and Optimal Control
\end{IEEEkeywords}

\section{INTRODUCTION}
\IEEEPARstart{W}{heels} are one of the major technological advances of humankind. In daily life, they enable us to move faster and more efficiently as compared to legged-based locomotion. The latter, however, is more versatile and offers the possibility to negotiate challenging environments, which is why combining both strategies into one system, would achieve the best of both worlds.

\begin{figure}[t]
    \centering
    \includegraphics[width=\columnwidth]{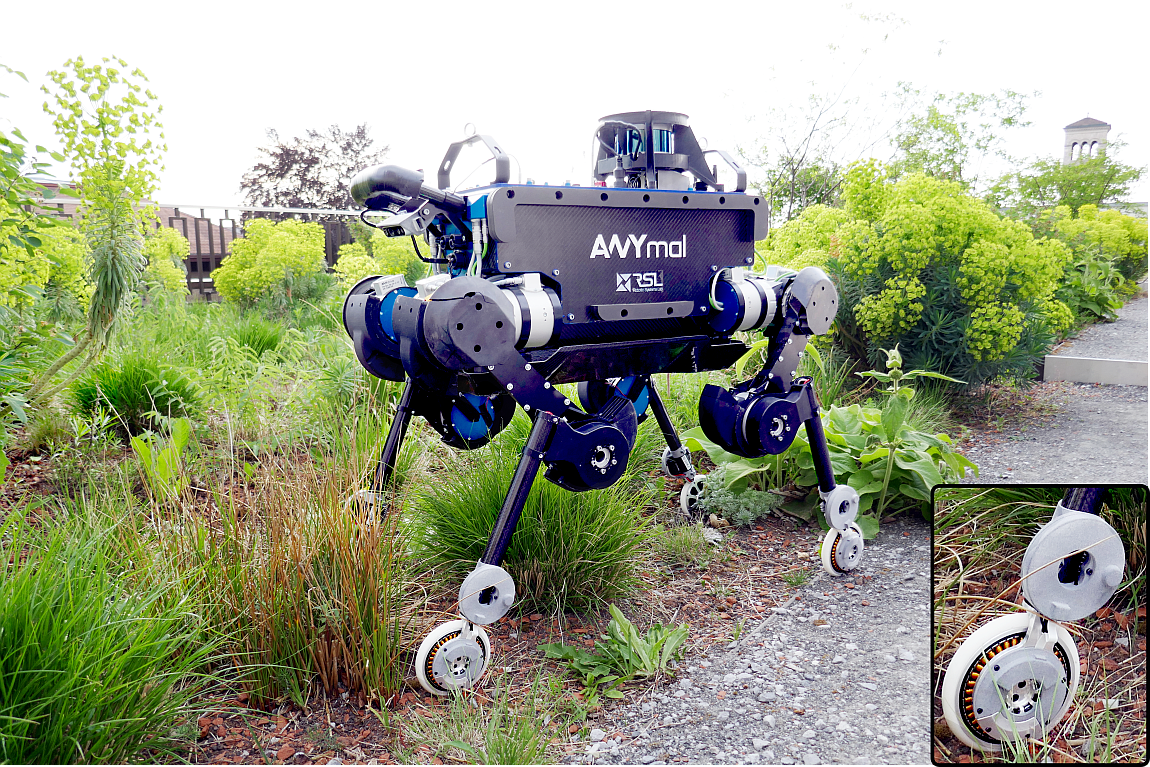}
     \caption{The fully torque-controlled quadrupedal robot ANYmal~\cite{hutter2016anymal} is equipped with four non-steerable, torque-controlled wheels. Thus, the number of actuated joint coordinates $n_{\tau}$ and the number of joints $n_j$ are both equal to 16. A video demonstrating the results can be found at \href{https://youtu.be/nGLUsyx9Vvc}{https://youtu.be/nGLUsyx9Vvc}.}
    \label{fig:anymal_on_wheels}
    \vspace{-0.5cm}
\end{figure}

While most of the advances towards autonomous mobile robots either focus on pure walking or driving, this paper shows how to plan and control trajectories for wheeled-legged robots as depicted in Fig.~\ref{fig:anymal_on_wheels} to achieve dynamic locomotion. We believe that such kinds of systems offer the solution for many robotic tasks as described in~\cite{bellicoso2018jfr}, e.g., rapid exploration, payload delivery, search and rescue, and industrial inspection.
\subsection{Related Work}
Recent years have shown an active research area focusing on the combination of wheeled and legged locomotion. Most wheeled-legged robots, such as~\cite{reid2016actively,giordano2009kinematic,cordes2014active,giftthaler2017efficient,suzumura2014real,grand2010motion}, behave like an active suspension system while driving and do not use their legs as a locomotion alternative to the wheels. While these wheeled-legged robots are using a kinematic approach to generate velocity commands for the wheels, there has been some promising research incorporating the whole-body dynamics of the robot to generate torque commands for each of the joints, including the wheels.

The authors in~\cite{sentis2013implementation} show a prioritized whole-body compliant control framework that generates motor torques for the upper body of a humanoid robot attached to a wheeled base. The equations of motion, including the nonholonomic constraints, are also incorporated into the control structure of a two-wheeled mobile robot~\cite{jeong2008wheeled}. \emph{Justin}~\cite{dietrich2016whole}, a wheeled humanoid robot, creates torque commands for each of the wheels using an admittance-based velocity controller. Each of these wheeled platforms, however, is not able to step due to the missing legs, and as such, the robots are only performing wheeled locomotion.

In contrast, \emph{DRC-HUBO+}~\cite{lim2017robot} is a wheeled humanoid robot which is able to switch between a walking and a driving configuration. While driving, the robot is in a crouched position, and as such, the legs are not used for locomotion or balancing.

\emph{Momaro}~\cite{klamt2017anytime}, on the other hand, shows driving and stepping without changing its configuration. This wheeled quadrupedal robot uses a kinematic approach to drive and to overcome obstacles like stairs and steps. Recently, the \emph{Centauro} robot~\cite{klamt2013Supervised,laurenzi2018quadrupedal,kamedula2018on} showed similar results over stepping stones, steps and first attempts to overcome stairs, while performing only slow static maneuvers.

There is a clear research gap for wheeled-legged robots. Most of the robots using actuated wheels are not taking into account the dynamic model of the whole-body including the wheels. The lack of these model properties hinders these robots from performing dynamic locomotion during walking and driving. In particular, a wheeled-legged robot produces reaction forces between its wheels and the terrain to generate its motion. The switching of the legs' contact state, the additional \ac{DOF} along the rolling direction of each wheel, and the reaction forces, all need to be accounted for in order to reveal the potential of wheeled-legged robots compared to traditional legged systems. In addition, torque control for the wheels is only explored for some slowly moving wheeled mobile platforms. Without force or torque control, the friction constraints related to the no-slip condition cannot be fulfilled, and locomotion is not robust against unknown terrain irregularities. Research areas in traditional legged locomotion~\cite{bellicoso2017dynamic,hutter2016anymal,pratt1997virtual,gehring2016practice,bledt2017policy,park2017high,winkler2018gait}, however, offer solutions to bridge these gaps. To this end, the work in~\cite{geilinger2018skaterbots} shows a generic approach to generate motions for wheeled-legged robots. Due to the formulation of the \ac{NLP} problem, the computation is too slow to execute in a receding horizon fashion, which is needed for robust execution under external disturbances. Moreover, the same authors verified their \ac{NLP} algorithm on rather small robots. 

So far, Boston Dynamics' wheeled bipedal robot \emph{Handle}~\cite{handle} is the only solution that demonstrated dynamic motions to overcome high obstacles while showing adaptability against terrain irregularities. Due to the missing publications on Handle, there is no knowledge about Boston Dynamics' locomotion framework.
\subsection{Contribution}
This paper shows dynamic locomotion for wheeled quadrupedal robots which combine the mobility of legs with the efficiency of driving. Our main contribution is a whole-body motion control and planning framework which takes into account the additional degrees of freedom introduced by torque-controlled wheels. The motion planner relies on an online \ac{ZMP}~\cite{vukobratovic2004zero} based optimization which continuously updates reference trajectories for the free-floating base and the wheels in a receding horizon fashion. These optimized motion plans are tracked by a hierarchical \ac{WBC} which takes into account the nonholonomic constraints introduced by the wheels. In contrast to other wheeled-legged robots, all joints including the wheels are torque controlled. To the best of our knowledge, this work shows for the first time dynamic and hybrid locomotion over flat, inclined and rough terrain for a wheeled quadrupedal robot. Moreover, we show how the same whole-body motion controller and planner are applied to driving and walking without changing any of the principles of dynamics and balance. 
\section{MODELLING OF WHEELED-LEGGED ROBOTS}
\label{sec:modeling_of_wheeled_legged_robots}
\begin{figure}[b]
    \vspace{-0.5cm}
    \centering
    \includegraphics[width=\columnwidth]{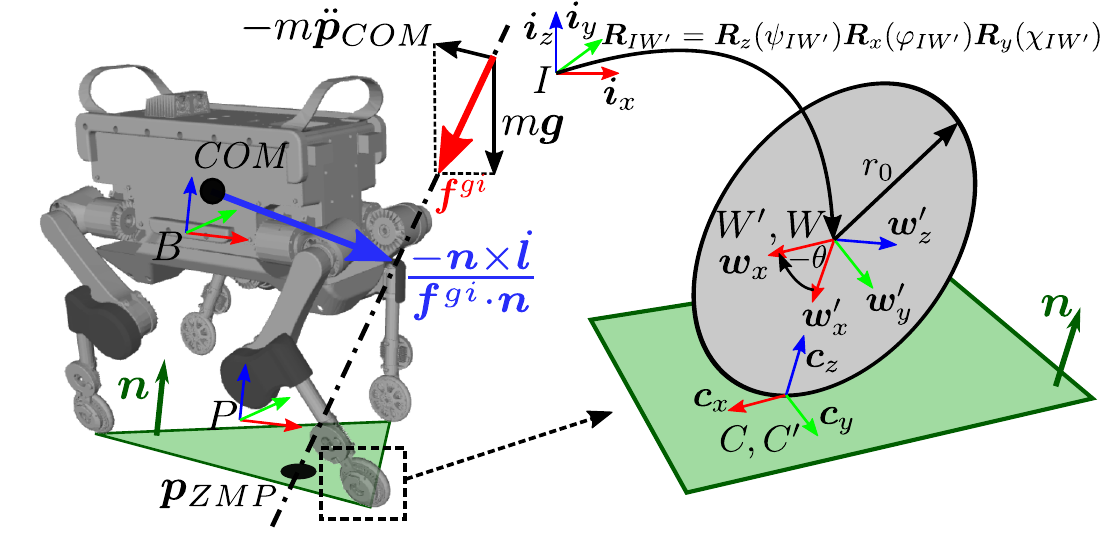}
    \caption{The figure illustrates a sketch of the wheeled quadrupedal robot ANYmal and the wheel model used to derive the rolling constraint (\ref{eq:rolling_constraint}). 
    \textbf{Left figure:} As discussed in~\cite{bellicoso2017dynamic}, we define a plan frame $P$ which is used as a reference frame in our motion planner. The red and blue arrows visualize the gravito-inertial wrench of the 3D ZMP model described in Section~\ref{sec:3d_zmp_inequality_constraint}.
    \textbf{Right figure:} We differentiate between the leg-fixed and wheel-fixed coordinate frames at the wheel. The \emph{leg-fixed} wheel frame $W'$ and contact frame $C'$ do not depend on the joint angle $\theta$ of the wheel. In contrast, the \emph{wheel-fixed} wheel frame $W'$ and contact frame $C'$ depend on the joint angle $\theta$ of the wheel.
    Both contact frames are aligned with the local estimation of the terrain normal $\bm{n}$ and the rolling direction $\bm{c}_x$ of the wheel.}
    \label{fig:zmp_and_frames}
\end{figure}

We first recall basic definitions of the kinematics and dynamics of robotic systems. Similar to walking robots~\cite{bellicoso2017dynamic}, a wheeled-legged robot is modeled as a \emph{free-floating base} $B$ to which the legs including the wheels as end-effectors are attached. Given a fixed \emph{inertial frame} $I$ (see Fig.~\ref{fig:zmp_and_frames}), the position from frame $I$ to $B$ \ac{w.r.t.} frame $I$ and the orientation of frame $B$ \ac{w.r.t.} frame $I$ are described by $_{I}\bm{r}_{IB} \in \mathbb{R}^{3}$ and a Hamiltonian unit quaternion $\bm{q}_{IB}$ . The generalized coordinate vector $\bm{q}$ and the generalized velocity vector $\bm{u}$ are given by
\begin{equation} \label{eq:generalized_q_and_u}
\bm{q} = 
\begin{bmatrix}
_{I}\bm{r}_{IB} \\ \bm{q}_{IB} \\ \bm{q}_{j}
\end{bmatrix} \in SE(3) \times \mathbb{R}^{n_j},
\bm{u} = 
\begin{bmatrix}
_{I}\bm{v}_{B} \\ _{B}\bm{\omega}_{IB} \\ \dot{\bm{q}}_{j}
\end{bmatrix} \in \mathbb{R}^{n_u},
\end{equation}
where $\bm{q}_{j} \in \mathbb{R}^{n_j}$ is the vector of joint coordinates, with $n_j$ the number of joint coordinates, $n_u = 6 + n_j$ is the number of generalized velocity coordinates, $_{I}\bm{v}_{B} \in \mathbb{R}^{3}$ is the linear velocity of frame $B$ \ac{w.r.t.} frame $I$, and $_{B}\bm{\omega}_{IB} \in \mathbb{R}^{3}$ is the angular velocity from frame $I$ to $B$ \ac{w.r.t.} frame $B$. With this convention, the equations of motion for wheeled-legged robots are defined by
\begin{equation} \label{eq:equation_of_motion}
\bm{M}(\bm{q}) \dot{\bm{u}} + \bm{h}(\bm{q},\bm{u}) 
= \bm{S}^T \bm{\tau} + \bm{J}_S^T \bm{\lambda},
\end{equation}
where $\bm{M}(\bm{q}) \in \mathbb{R}^{n_u \times n_u}$ is the mass matrix, $\bm{h}(\bm{q},\bm{u}) \in \mathbb{R}^{n_u}$ is the vector of Coriolis, centrifugal and gravity terms, $\bm{\tau} \in \mathbb{R}^{n_\tau}$ is the generalized torque vector acting in the direction of the generalized coordinate vector, with $n_{\tau}$ the number of actuated joint coordinates, $\bm{J}_S = [\bm{J}_{C_1}^T \ {\dots} \ \bm{J}_{C_{n_c}}^T]^T \in \mathbb{R}^{3n_c \times n_u}$ is the support Jacobian, with $n_c$ the number of limbs in contact, and $\bm{\lambda} \in \mathbb{R}^{3n_c}$ is the vector of constraint forces. The transpose of the selection matrix $\bm{S} = [\bm{0}_{n_{\tau} \times n_{u}-n_{\tau}} \ \mathbb{I}_{n_{\tau} \times n_{\tau}}]$ maps the generalized torque vector $\bm{\tau}$ to the space of generalized forces.
\subsection{Nonholonomic Rolling Constraint}
\label{sec:rolling_constraint}
In contrast to point contacts, the acceleration of the wheel-fixed contact point\footnote{In contrast to the wheel-fixed contact point $C_i$, the leg-fixed contact point $C'_i$ does not need to have zero velocity.} $C_i$ of the $i$-th leg does not equal zero, i.e., $_{I}\ddot{\bm{r}}_{IC_i} = \bm{J}_{C_i} \dot{\bm{u}} + \dot{\bm{J}}_{C_i} \bm{u} \neq \bm{0}$. Given the wheel model in Fig.~\ref{fig:zmp_and_frames}, it can be shown that the resulting contact acceleration of a wheel is defined by
\begin{equation} \label{eq:rolling_constraint}
\begin{aligned}
& _{I}\ddot{\bm{r}}_{IC_i}
= \bm{J}_{C_i} \dot{\bm{u}} + \dot{\bm{J}}_{C_i} \bm{u}
= \\
& \bm{R}_{IW_i}
\begin{bmatrix}
0 \\
- r_0 \dot{\psi}_{IW'_i} \cos(\varphi_{IW'_i})(\dot{\chi}_{IW'_i}+\dot{\theta}) \\
r_0 (\dot{\chi}_{IW'_i}+\dot{\theta})(\dot{\chi}_{IW'_i}+\dot{\theta}+ \dot{\psi}_{IW'_i} \sin(\varphi_{IW'_i}))
\end{bmatrix},
\end{aligned}
\end{equation}
where $\bm{R}_{IW_i} \in SO(3)$ represents the rotation matrix that projects
the components of a vector from the \emph{wheel frame} $W_i$ to the inertial frame $I$, $r_0$ is the wheel radius, and $\theta_i$ is the joint angle of the wheel. Using an intrinsic $z-x'-y''$ Euler parameterization, the yaw, roll, and pitch angle of the \emph{wheel fixed frame} $W'_i$ \ac{w.r.t.} the inertial frame $I$ are given by $\psi_{IW'_i}$, $\varphi_{IW'_i}$, and $\chi_{IW'_i}$, respectively. 

By setting $\varphi_{IW'_i} \equiv 0$ and  $\psi_{IW'_i} \equiv 0$, we obtain the acceleration for the planar case, i.e, $_{I}\ddot{\bm{r}}_{IC_i} = \bm{R}_{IW_i} [0\ 0\ r_0 (\dot{\chi}_{IW'_i}+\dot{\theta})^2]^T$, which is equal to the centripetal acceleration.
\subsection{Terrain and Contact Point Estimation}
\label{sec:terrain_and_contact_estimation}
The robot is blindly locomoting on a terrain locally modeled by a three-dimensional plane. First, the terrain normal is estimated by fitting a plane through the most recent contact locations of the wheel frame $W$ in Fig.~\ref{fig:zmp_and_frames} using a least-squares method as described in~\cite{gehring2016practice}. Given the resulting terrain normal $\bm{n}$, the estimated plane is moved along the terrain normal to the contact position $C$ as illustrated in Fig.~\ref{fig:zmp_and_frames}, i.e., the terrain plane is shifted by $\pi_{{W_{x,z}}}(-\bm{n})R/ |\pi_{{W^{x,z}}}(-\bm{n})|$, where $\pi_{{W_{x,z}}}(-\bm{n})$ is the projection of the negative normal vector $\bm{n}$ onto the plane spanned by $\bm{w}_{x}$ and $\bm{w}_{z}$. Finally, the plane through the contact points represents the estimated terrain plane used for control and planning.

The \emph{leg-fixed contact frame}\footnote{The leg-fixed contact frame $C'_i$ is defined as a point \ac{w.r.t.} the leg-fixed wheel frame $W'_i$. It follows that the Jacobian $\bm{J}_{C'_i}$ does not depend on the joint angle $\theta_i$ of the $i$-th wheel.} $C'_i$ and \emph{wheel-fixed contact frame}\footnote{The wheel-fixed contact frame $C_i$ is defined as a point \ac{w.r.t.} the wheel frame $W_i$. It follows that the Jacobian $\bm{J}_{C_i}$ depends on the joint angle $\theta_i$ of the $i$-th wheel.} $C_i$ of each leg $i$ are introduced to simplify the convention of the motion controller and planner. As illustrated in Fig.~\ref{fig:zmp_and_frames}, both contact frames are defined to lie at the intersection of the wheel plane with the estimated terrain plane. The contact frame's $z$-axis is aligned with the estimated terrain normal and its $x$-axis is perpendicular to the estimated terrain normal and aligned with the rolling direction\footnote{The rolling direction of the wheel is computed by $\bm{c}_x=\bm{w}_y \times \bm{n}/|\bm{w}_y \times \bm{n}|$.} $\bm{c}_x$ of the wheel.

As discussed in earlier works~\cite{bellicoso2017dynamic}, the motion plans in Section~\ref{sec:3d_zmp} are computed in the \emph{plan frame} $P$ whose $z$-axis is aligned with the estimated terrain normal and whose $x$-axis is perpendicular to the estimated terrain normal and aligned with the heading direction of the robot. As depicted in Fig.~\ref{fig:zmp_and_frames}, the plan frame is located at the footprint center projected onto the local terrain along the terrain normal.
\section{MOTION PLANNING}
\label{sec:3d_zmp}
The dynamic model of a wheeled-legged robot (\ref{eq:equation_of_motion}) includes significant nonlinearities to be handled by the motion planner. Due to this complexity, the optimization problem becomes prone to local minima
and it can be demanding to solve in real-time on-board~\cite{bledt2017policy}.
To overcome these challenges, our approach breaks down the whole-body planning problem into \ac{COM} and foothold motion optimization~\cite{bellicoso2017dynamic,bellicoso2017dynamicAndWholeBody}. We simplify the system dynamics to a \ac{ZMP} model for motion planning of the \ac{COM}. The reference footholds for each leg are obtained by solving a separate optimization problem.

\begin{figure}[b!]
    \vspace{-0.5cm}
    \centering
    \includegraphics[width=\columnwidth]{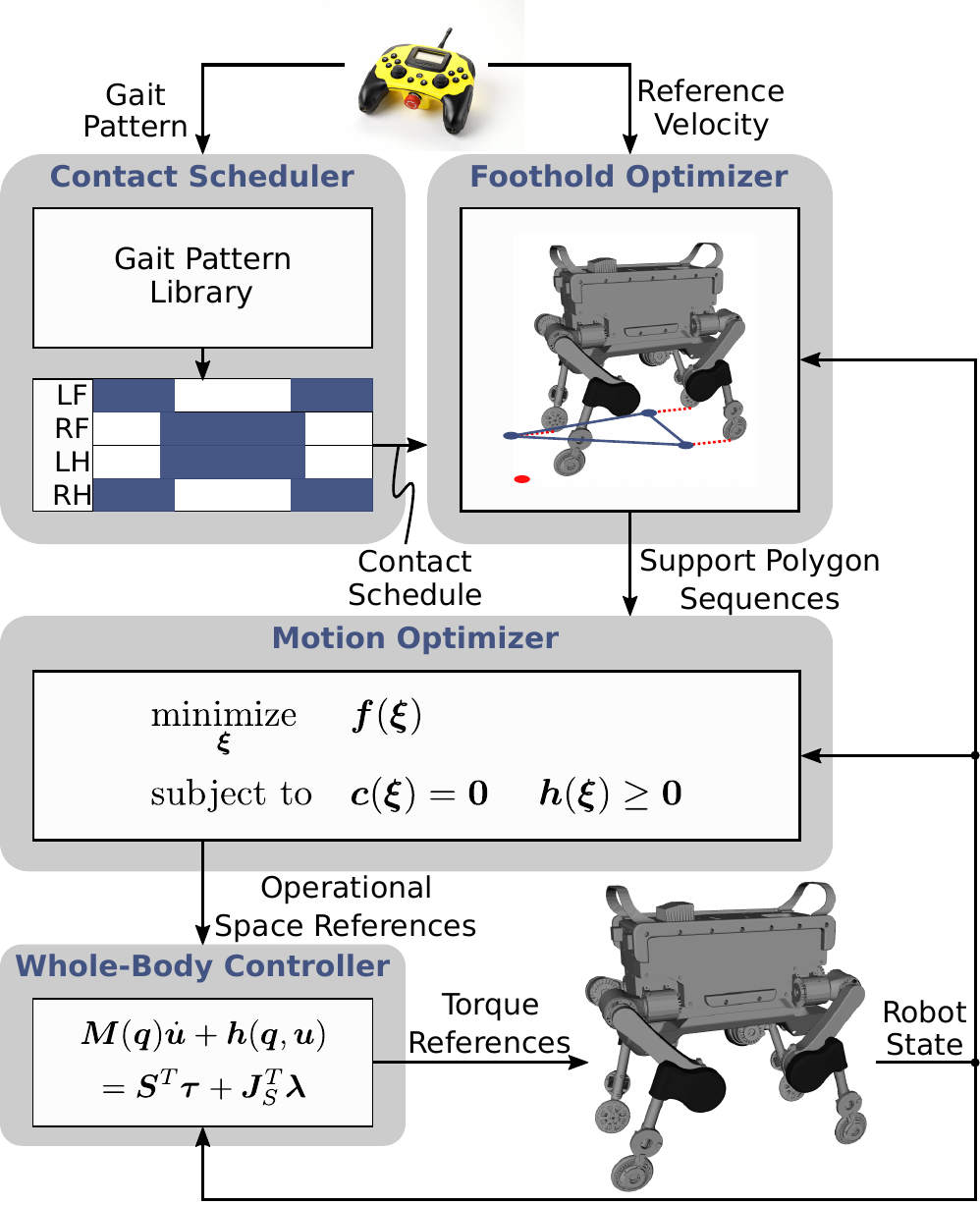}
    \caption{The motion planner is based on a 3D \ac{ZMP} approach which takes into account the support polygon sequence and the state of the robot. The hierarchical \ac{WBC} which optimizes the whole-body accelerations and contact forces tracks the operational space references. Finally, torque references are sent to the robot.}
    \label{fig:approach}
\end{figure}

Fig.~\ref{fig:approach} gives an overview of the entire whole-body motion control and planning framework. The foothold optimizer, motion optimizer, and \ac{WBC} modules are solving separate optimization problems in parallel such that there is no interruption between them~\cite{bellicoso2017dynamic}. We generate all motions \ac{w.r.t.} the plan frame $P$ introduced in Section~\ref{sec:terrain_and_contact_estimation}. In the following, we describe each module of the motion planner.
\subsection{Contact Scheduler}
The contact schedule defines periodic sequences of lift-off and touch-down events for each leg. Based on a gait pattern library, each gait predefines the timings for each leg over a stride, e.g., the contact scheduler block in Fig.~\ref{fig:approach} illustrates the gait pattern for a trotting gait. With this formulation, driving is defined by a gait pattern where each leg is scheduled to stay in contact, and no lift-off events are set.
\subsection{Foothold Optimizer}
Given a reference in terms of linear velocity $\bm{v}_{B}^{ref} = [v_{B,x}^{ref}\ v_{B,y}^{ref}\ 0]^T$ and angular velocity $\bm{\omega}_{B}^{ref}=[0\ 0\ \omega_{B,z}^{ref}]^T$ of the base, and the contact schedule, desired footholds\footnote{A foothold is the contact position $C$ of a grounded leg.} are generated for each leg. Based on the contact schedule and footholds, a sequence of support polygons are generated, where each polygon is defined by the convex hull of expected footholds, e.g., the green polygon in Fig.~\ref{fig:zmp_and_frames}, as well as its time duration in seconds.  

While walking, we formulate a \ac{QP} problem which optimizes over the $x$ and $y$ coordinates of each foothold~\cite{bellicoso2017dynamic}. Costs, which are added to the \ac{QP} problem, penalize the distance between the optimized foothold locations and different contributions to the computation of the footholds. We assign default foothold positions which define the standing configuration of the robot. Footholds are projected using the high-level reference velocity and assuming constant reference velocity throughout the optimization horizon. To ensure smoothness of the footholds, we penalize the deviation from previously computed footholds. Finally, we rely on an inverted pendulum model to stabilize the robot's motion~\cite{gehring2016practice}. Inequality constraints are added to avoid collisions of the feet and to respect the maximum kinematic extension of each leg. Given the previous stance foot position and the optimized foothold, a swing trajectory for each leg is generated by fitting a spline between both.

Traditional legged locomotion is based on the constraint that the leg-fixed contact point $C'$ remains stationary when in contact with the environment. In contrast, wheeled-legged robots are capable of executing trajectories along the rolling direction $\bm{c}_x$ of the wheel. This can be seen as a moving foothold. While driving, the desired leg-fixed contact position $_{I}\bm{r}_{IC'_i}^d \in \mathbb{R}^{3}$, velocity $_{I}\dot{\bm{r}}_{IC'_i}^d \in \mathbb{R}^{3}$ and acceleration $_{I}\ddot{\bm{r}}_{IC'_i}^d \in \mathbb{R}^{3}$ of leg $i$ are computed based on the reference velocities $\bm{v}_{B}^{ref}$ and $\bm{\omega}_{B}^{ref}$ of the base and the state of the robot.
\subsection{Motion Optimizer}
\label{sec:motion_optimization}
The motion optimizer generates operational space references for the $x$, $y$ and $z$ coordinates of the whole-body \ac{COM} given the support polygon sequence and the robot state~\cite{bellicoso2017dynamic}. The resulting nonlinear optimization framework is described in the following sections.
\subsubsection{Motion plan parameterization}
\label{sec:motion_plan_parametrization}
The $x$, $y$, and $z$ coordinates of the \ac{COM} trajectory are parametrized as a sequence of quintic splines~\cite{bellicoso2017dynamic}, i.e., the position, velocity and acceleration of the \ac{COM} are given by $\bm{p}_{COM}=\bm{T}(t)\bm{\alpha}_k \in \mathbb{R}^3$, $\dot{\bm{p}}_{COM}=\dot{\bm{T}}(t)\bm{\alpha}_k \in \mathbb{R}^3$, and $\ddot{\bm{p}}_{COM}=\ddot{\bm{T}}(t)\bm{\alpha}_k \in \mathbb{R}^3$, with $\bm{T}(t) = diag(\bm{\eta}^T(t), \bm{\eta}^T(t), \bm{\eta}^T(t)) \in \mathbb{R}^{3 \times 18}$, $\bm{\eta}^T(t) = [t^5 \  t^4 \ t^3 \ t^2 \ t \ 1]$, $t \in [\bar{t},\bar{t}+\Delta t_k]$, where $\bar{t}$ is the sum of time durations of all previous splines, and $\Delta t_k$ is the time duration of the $k$-th spline. All coefficients of spline $i$ are stored in $\bm{\alpha}_k = [{\bm{\alpha}_k^x}^T \ {\bm{\alpha}_k^y}^T \ {\bm{\alpha}_k^z}^T ]^T \in \mathbb{R}^{18}$. Finally, we solve for the vector of optimization parameters which is obtained by stacking together all spline coefficient vectors $\bm{\alpha}_k$.
\subsubsection{Optimization problem}
\label{sec:sqp}
The motion optimization problem is expressed as a nonlinear optimization
problem with objective $\bm{f}(\bm{\xi})$, equality constraints $\bm{c}(\bm{\xi})$, and inequality constraints $\bm{h}(\bm{\xi})$. The problem is described by
\begin{equation} \label{eq:sqp}
\begin{aligned}
& \underset{\bm{\xi}}{\text{minimize}}
& & \bm{f}(\bm{\xi}) \\
& \text{subject to}
& & \bm{c}(\bm{\xi}) = \bm{0}, \ \ \ \ \bm{h}(\bm{\xi}) \geq \bm{0},
\end{aligned}
\end{equation}
where $\bm{\xi}$ is the vector of optimization variables given in Section~\ref{sec:motion_plan_parametrization}, i.e., optimal spline coefficients are computed. A \ac{SQP} algorithm~\cite{boggs1995sequential} is used to solve (\ref{eq:sqp}) continuously over a time horizon of $\tau$ seconds. Table~\ref{table:optimization_problem} summarizes each objective and constraint used in this work.
%

\begin{table}[b!]
    \vspace{-0.5cm}
    \caption{The table lists the costs and constraints of the motion optimization problem based on~\cite{bellicoso2017dynamic}.}
    \label{table:optimization_problem}
    \begin{center}
    \begin{tabular}{ccc}
    \toprule
    Type &Task &Purpose \\\midrule
    Objective & \begin{tabular}{@{}c@{}}Minimize\\ \ac{COM} acceleration\end{tabular} &Smooth motions \\
    Objective & \begin{tabular}{@{}c@{}}Minimize deviation to\\ previous solution $\bm{\xi}_{prev}$\end{tabular} & Smooth motions \\
    Objective & \begin{tabular}{@{}c@{}c@{}}Track a high-level \\ reference trajectory $\bm{\pi}$ \\ (path regularizer) $\forall \,\bm{\xi}$ \end{tabular} & Reference tracking \\ \\
    \begin{tabular}{@{}c@{}}Soft constraint\\ (lin.-quad.)\end{tabular} &  \begin{tabular}{@{}c@{}c@{}}Minimize deviation to\\ initial \& final \\ conditions  $\forall \,\bm{\xi}$\end{tabular} & \begin{tabular}{@{}c@{}}Disturbance rejection \\ \& reference tracking\end{tabular} \\
    \begin{tabular}{@{}c@{}}Soft constraint\\ (lin.-quad.)\end{tabular} & Limit overshoots $\forall \,\bm{\xi}^z $ & \begin{tabular}{@{}c@{}}Avoid kinematic \\ limits of legs\end{tabular} \\ \\
    \begin{tabular}{@{}c@{}}Constraint\\ (lin. eq.)\end{tabular} & \begin{tabular}{@{}c@{}c@{}}Junction constraints $\forall$\\pairs of adjacent splines\\ $k$, $k+1$ $\forall \,\bm{\xi}$\end{tabular} & Continuity \\
    \begin{tabular}{@{}c@{}}Constraint\\ (lin. ineq.)\end{tabular} & \begin{tabular}{@{}c@{}}Push Contact\\ Constraints\end{tabular} &\begin{tabular}{@{}c@{}}Legs can only\\ push the ground\end{tabular} \\ \\
    \begin{tabular}{@{}c@{}}Constraint\\ (nonlin. ineq.)\end{tabular} & \ac{ZMP} criterion & Stability \\
    \begin{tabular}{@{}c@{}}Soft constraint\\ (nonlin.)\end{tabular} & \begin{tabular}{@{}c@{}}Soften initial\\ \ac{ZMP} constraints \end{tabular} & Relaxation
    \\\bottomrule
    \end{tabular}
    \end{center}
\end{table}
\subsubsection{ZMP inequality constraint}
\label{sec:3d_zmp_inequality_constraint}
To ensure dynamic stability of the planned motions, an inequality constraint on the \ac{ZMP} position $\bm{p}_{ZMP} \in \mathbb{R}^3$ is included in the motion optimization, where $\bm{p}_{ZMP} = \bm{n} \times \bm{m}_O^{gi}/ (\bm{n}^T \bm{f}^{gi})$~\cite{sardain2004forces}. Here, $\bm{m}_O^{gi} \in \mathbb{R}^3$ and $\bm{f}^{gi} \in \mathbb{R}^3$ are the components of the \emph{gravito-inertial} wrench~\cite{caron2017zmp}, with $\bm{m}_O^{gi}=m \cdot \bm{p}_{COM} \times (\bm{g}-\ddot{\bm{p}}_{COM}) - \dot{\bm{l}}$ and $\bm{f}^{gi} = m \cdot (\bm{g} - \ddot{\bm{p}}_{COM})$, where $m$ is the mass of the robot, $\bm{l} \in \mathbb{R}^3$ is the angular momentum of the \ac{COM}, and $\bm{g} \in \mathbb{R}^3$ is the gravity vector. Fig.~\ref{fig:zmp_and_frames} shows a sketch of the gravito-inertial wrench acting at the \ac{COM}. As in~\cite{bellicoso2017dynamic}, we assume that $\dot{\bm{l}}=\bm{0}$.

As illustrated in Fig.~\ref{fig:zmp_and_frames}, the \ac{ZMP} position $\bm{p}_{ZMP}$ is constrained to lie inside the support polygon. This stability criterion is formulated as a nonlinear inequality constraint given by~\cite{bellicoso2017dynamic}
\begin{equation} \label{eq:zmp_constraint}
\begin{bmatrix}
p & q & 0
\end{bmatrix}
 \bm{p}_{ZMP}+r \geq 0,
\end{equation}
where $\bm{d}=[p \ q \ r]^T$ are the coefficients of the line that goes through the edge of a support polygon. 
\subsubsection{Deformation of support polygons while driving}
In contrast to point feet, the contact locations, and therefore footholds, are not stationary while driving. The support polygon sequence which is needed to fulfill the inequality constraint in (\ref{eq:zmp_constraint}) is deformed over time. For this purpose, we assume that the number of edges stays constant and therefore, one spline is sufficient to describe the motion of the \ac{COM}. 

First, the expected foothold position for the optimization horizon $\tau$ is computed as a function of the reference velocities $\bm{v}_{B}^{ref}$ and $\bm{\omega}_{B}^{ref}$. The reference velocities are assumed to be constant over the optimization horizon. Using the time-integrated Rodriguez's formula, the expected foothold position $\bm{p}_{\tau,i} \in \mathbb{R}^3$ of leg $i$ is computed by
\begin{equation} \label{eq:rodriguez_formula}
\begin{aligned}
&\bm{p}_{\tau,i} = \bm{p}_{0,i} + \bm{R}(\tau \bm{\omega}_{B}^{ref})\\ 
& \frac{1}{\omega_{B,z}^{ref}}
\begin{bmatrix}
\sin(\omega_{B,z}^{ref} \tau) & -1+\cos(\omega_{B,z}^{ref} \tau)& 0 \\
1-\cos(\omega_{B,z}^{ref} \tau) & \sin(\omega_{B,z}^{ref} \tau) & 0 \\
0 & 0 & 0
\end{bmatrix} \bm{v}_{B}^{ref},
\end{aligned}
\end{equation}
where $\bm{p}_{0,i} \in \mathbb{R}^3$ is the current foothold position. If $\omega_{B,z}^{ref} \approx 0$, the solution becomes $\bm{p}_{\tau,i} = \bm{p}_{0,i} + \tau \bm{v}_{B}^{ref}$.

Given the coefficients which describe an edge that belongs to the current and expected support polygon, i.e., $\bm{d}_{0} \in \mathbb{R}^3$ and $\bm{d}_{\tau} \in \mathbb{R}^3$, the deformed edge coefficient vector $\bm{d}_k(t)$ at time $t$ is computed by interpolating $\bm{d}_{0}$ and $\bm{d}_{\tau}$, i.e.,
\begin{equation} \label{eq:support_polygon_interpolation}
\bm{d}(t) = (1-\frac{t-\bar{t}}{\tau}) \bm{d}_{\tau}  + \frac{t-\bar{t}}{\tau} \bm{d}_{0}.
\end{equation}
\section{WHOLE-BODY CONTROLLER}
The operational space reference trajectories of the \ac{COM} and wheels are tracked by a \ac{WBC} which is based on the \ac{HO} framework described in~\cite{bellicoso2017dynamic,bellicoso2017dynamicAndWholeBody}. We compute optimal generalized accelerations $\dot{\bm{u}}^*$ and contact forces $\bm{\lambda}^*$ which are collected in the vector of optimization variables $\bm{\xi}^*=[\dot{\bm{u}}^{*T} \ \bm{\lambda}^{*T}]^T \in \mathbb{R}^{n_u + 3 n_c}$, where all symbols are introduced in Section~\ref{sec:modeling_of_wheeled_legged_robots}.

The \ac{WBC} is formulated as a cascade of \ac{QP} problems composed of linear equality and inequality tasks, which are solved in a strict prioritized order~\cite{bellicoso2016perception}. A task $T_p$ with priority $p$ is defined by
\begin{equation}\label{eq:wbc_task}
T_p :
\begin{cases}
\bm{W}_{eq,p}(\bm{A}_p \bm{\xi} - \bm{b}_p) = \bm{0} \\
\bm{W}_{ineq,p}(\bm{D}_p \bm{\xi} - \bm{f}_p) \le \bm{0}
\end{cases},
\end{equation}
where the linear equality constraints are defined by $\bm{A}_p$ and $\bm{b}_p$, the linear inequality constraints are defined by $\bm{D}_p$ and $\bm{f}_p$, and the diagonal positive-definite matrices $\bm{W}_{eq,p}$ and $\bm{W}_{ineq,p}$ weigh tasks on the same priority level.
\begin{table}[t]
    \caption{The table lists the prioritized tasks (priority 1 is the highest) used in the \ac{WBC}. Bold tasks are tailored for wheeled-legged robots.}
    \label{table:tasks}
    \begin{center}
    \begin{tabular}{cl}
    \toprule
    Priority &Task \\\midrule
    1 & Floating base equations of motion \\
      & Torque limits and friction cone \\
      & \textbf{Nonholonomic rolling constraint} \\
    2 & \ac{COM} linear and angular motion tracking \\
      & \textbf{Swing leg motion tracking} \\
      & \textbf{Swing wheel rotation minimization} \\
      & \textbf{Ground leg motion tracking} \\
    3 & Contact force minimization
    \\\bottomrule
    \end{tabular}
    \end{center}
     \vspace{-0.5cm}
\end{table}
%
\begin{figure*}[b!]
    \centering
    \includegraphics[width=\textwidth]{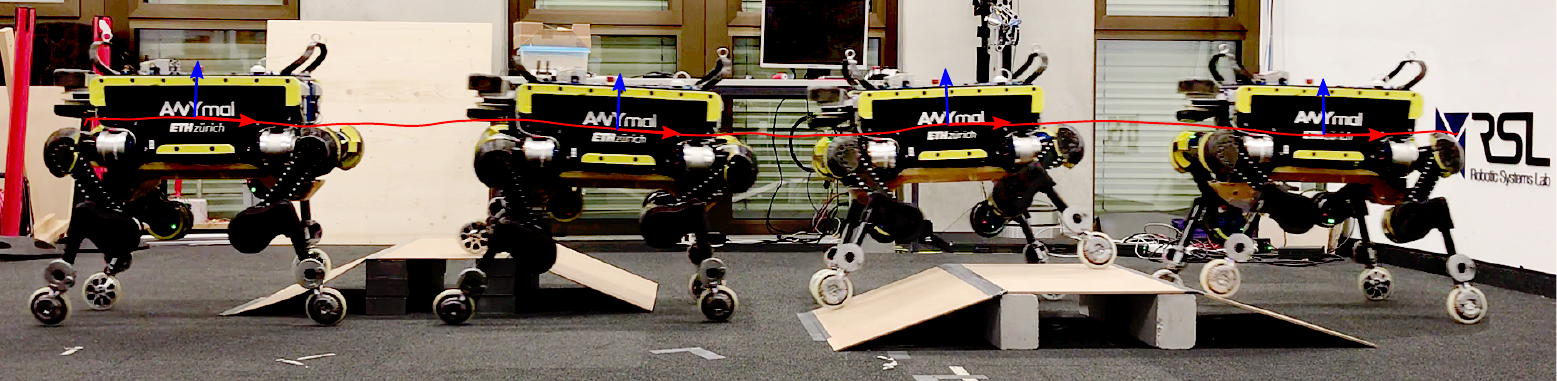}
    \caption{The robot ANYmal drives with a speed of 0.7\,m/s. over two inclines of a height of approximately 30\,\% of ANYmal's leg length and the red line depicts the \ac{COM} trajectory (Available at \href{https://youtu.be/nGLUsyx9Vvc?t=20}{https://youtu.be/nGLUsyx9Vvc?t=20}).}
    \label{fig:double_incline}
\end{figure*}
\subsection{Prioritized Tasks}
The highlighted tasks in Table~\ref{table:tasks} are specifically tailored for wheeled-legged robots, and the following sections describe each of these tasks in more detail. For the remaining tasks, we rely on the same implementation as used for traditional legged robots~\cite{bellicoso2017dynamicAndWholeBody}.

\emph{Floating base equations of motion}: The optimization vector $\bm{\xi}$ is constrained to be consistent with the system dynamics.

\emph{Torque limits and friction cone}: Inequality constraint tasks are added to the optimization problem to avoid that the computed torques exceed the minimum and maximum limit of each actuator. Similar, the contact forces $\bm{\lambda}$ need to lie inside the friction cone which is approximated by a friction pyramid and aligned with the normal vector $\bm{n}$ of the estimated contact surface shown in Fig.~\ref{fig:zmp_and_frames}.

\emph{Nonholonomic rolling constraint}: The solution found by the optimization needs to take into account the nonholonomic rolling constraint (\ref{eq:rolling_constraint}). This is expressed as an equality constraint given by
\begin{equation} \label{eq:wbc_rolling_constraint}
\begin{bmatrix}
\bm{J}_S & \bm{0}_{3 n_c \times 3 n_c}
\end{bmatrix}
\bm{\xi}_d =
-\dot{\bm{J}}_S \bm{u} +
\begin{bmatrix}
 _{I}\ddot{\bm{r}}_{IC_1}^T & {\dots} &  _{I}\ddot{\bm{r}}_{IC_{n_c}}^T
\end{bmatrix}^T,
\end{equation}
where the terms $_{I}\ddot{\bm{r}}_{IC_1} \ {\dots} \  _{I}\ddot{\bm{r}}_{IC_{n_c}}$ on the right side of the equation are the centripetal accelerations of each contact point $n_c$ derived in (\ref{eq:rolling_constraint}).

\emph{\ac{COM} linear and angular motion tracking}: Similar to the swing leg motion tracking task, the operational space references of the \ac{COM} are tracked by equality constraint tasks.

\emph{Swing leg motion tracking}: Given the operational space references of the wheels' contact points $_{P}\bm{r}_{IC'_i}^d$, $_{P}\dot{\bm{r}}_{IC'_i}^d$, and $_{P}\ddot{\bm{r}}_{IC'_i}^d$, the motion tracking task of each swing leg $i$ is formulated by
\begin{equation} \label{eq:wbc_swing_leg_tracking}
\begin{aligned}
\begin{bmatrix}
\bm{J}_{C'_i} & \bm{0}_{3 n_c \times 3 n_c}
\end{bmatrix}
\bm{\xi}_d =&
\bm{R}_{IP}({_{P}\ddot{\bm{r}}_{IC'_i}^d} +
\bm{K}_p ({_{P}\bm{r}_{IC'_i}^d} - {_{P}\bm{r}_{IC'_i}}) \\&
+ \bm{K}_d ({_{P}\dot{\bm{r}}_{IC'_i}^d} - {_{P}\dot{\bm{r}}_{IC'_i}}))
-\dot{\bm{J}}_{C'_i} \bm{u},
\end{aligned} 
\end{equation}
where $\bm{K}_p, \bm{K}_d \in \mathbb{R}^{3 \times 3}$ are diagonal positive definite matrices which define proportional and derivative gains. Note that all measured values, i.e., $\bm{J}_{C'_i}$, $_{P}\bm{r}_{IC'_i}$, and $_{P}\dot{\bm{r}}_{IC'_i}$, are independent of the wheel angle $\theta$ (as discussed in the footnotes of Section~\ref{sec:rolling_constraint}).

\emph{Swing wheel rotation minimization}: For each swing leg $i$, the wheel's rotation is damped by adding the task
\begin{equation} \label{eq:wb_swing_wheel}
\begin{bmatrix}
\bm{S}_{W_i} & \bm{0}_{3 n_c \times 3 n_c}
\end{bmatrix}
\bm{\xi}_d = -k_d \dot{\theta}_i,
\end{equation}
where $\bm{S}_{W_i} \in \mathbb{R}^{3n_c \times n_u}$ is a matrix which selects the row of $\bm{\xi}_d$ containing the wheel of leg $i$, $k_d$ is a derivative gain, and $\dot{\theta}_i$ is the wheel's rotational speed.

\emph{Ground leg motion tracking}: To track the desired motion of the grounded legs, we constrain the accelerations in the direction of the rolling direction $\bm{c}_x$. Given the operational space references of the wheels' contact points $_{P}\bm{r}_{IC'_i}^d$, $_{P}\dot{\bm{r}}_{IC'_i}^d$, and ${_{P}\ddot{\bm{r}}_{IC'_i}^d}$, the motion tracking task of each ground leg $i$ is formulated by
\begin{equation} \label{eq:wbc_ground_leg_tracking}
\begin{aligned}
& \pi_{\bm{c}_x}
(\begin{bmatrix}
\bm{J}_{C'_i} & \bm{0}_{3 n_c \times 3 n_c}
\end{bmatrix}
\bm{\xi}_d) =
\pi_{\bm{c}_x}(\bm{R}_{IP}({_{P}\ddot{\bm{r}}_{IC'_i}^d} \\&
+ \bm{K}_p ({_{P}\bm{r}_{IC'_i}^d} - {_{P}\bm{r}_{IC'_i}}) 
+ \bm{K}_d ({_{P}\dot{\bm{r}}_{IC'_i}^d} - {_{P}\dot{\bm{r}}_{IC'_i}}))
-\dot{\bm{J}}_{C'_i} \bm{u}),
\end{aligned} ,
\end{equation}
where $\pi_{\bm{c}_x}(\bm{a})$ is the projection of a vector $\bm{a}$ onto the vector $\bm{c}_x$.

\emph{Contact force minimization}: Finally, the contact forces $\bm{\lambda}$ are minimized to reduce slippage. 

\subsection{Torque Generation}

Given the optimal solution $\bm{\xi}^*$, the desired actuation torques $\bm{\tau}_d$, which are sent to the robot, are computed by
\begin{equation} \label{eq:reference_torques}
\bm{\tau}_d = \bm{M}_j(\bm{q}) \dot{\bm{u}}^* + \bm{h}_j(\bm{q},\bm{u}) - \bm{J}_{Sj}^T \bm{\lambda^*},
\end{equation}
where $\bm{M}_j(\bm{q})$, $\bm{h}_j(\bm{q},\bm{u})$, and $\bm{J}_{Sj}$ are the lower rows of the equations of motion in (\ref{eq:equation_of_motion}) relative to the actuated joints.
\section{EXPERIMENTAL RESULTS AND DISCUSSION}
To show the benefits and validity of our new approach, this section reports on experiments conducted on a real quadrupedal robot equipped with non-steerable, torque-controlled wheels. The robot is driven using external velocity inputs coming from a joystick. All computation was carried out by the PC (Intel i7-5600U, 2.6 - 3.2GHz, dual-core 64-bit) integrated into the robot. A video\footnote{\hbox{Available at \href{https://youtu.be/nGLUsyx9Vvc}{https://youtu.be/nGLUsyx9Vvc}}} showing the results accompanies this paper. 

The \ac{WBC} runs together with state estimation in a 400\,Hz loop. A novel state estimation algorithm based on~\cite{bloesch2018two} is used to generate an estimation of the robot's position, velocity, and orientation \ac{w.r.t.} an inertial coordinate frame. Similar to~\cite{bloesch2013stateconsistent}, we fuse data from an \ac{IMU} as well as the kinematic measurements from each actuator (including the wheels) to acquire a fast state estimation of the robot. The open-source Rigid Body Dynamics Library~\cite{rbdl} (RBDL) is used for modeling and computation of kinematics and dynamics based on the algorithms described in~\cite{featherstone2014rigid}. We use a custom \ac{SQP} algorithm to solve the nonlinear optimization problem in Section~\ref{sec:sqp}, which solves the nonlinear problem by iterating through a sequence of \ac{QP} problems. Each \ac{QP} problem is solved using QuadProg++~\cite{quadProg} which uses the Goldfarb-Idnani active-set method~\cite{goldfarb1983numerically}. Depending on the gait, the motion optimization in Section~\ref{sec:motion_optimization} runs between 100 and 200\,Hz.
\subsection{Indoor Environment: Flat Terrain}
\begin{figure}[b!]
    \centering
     \vspace{-0.5cm}
    \includegraphics[width=\columnwidth]{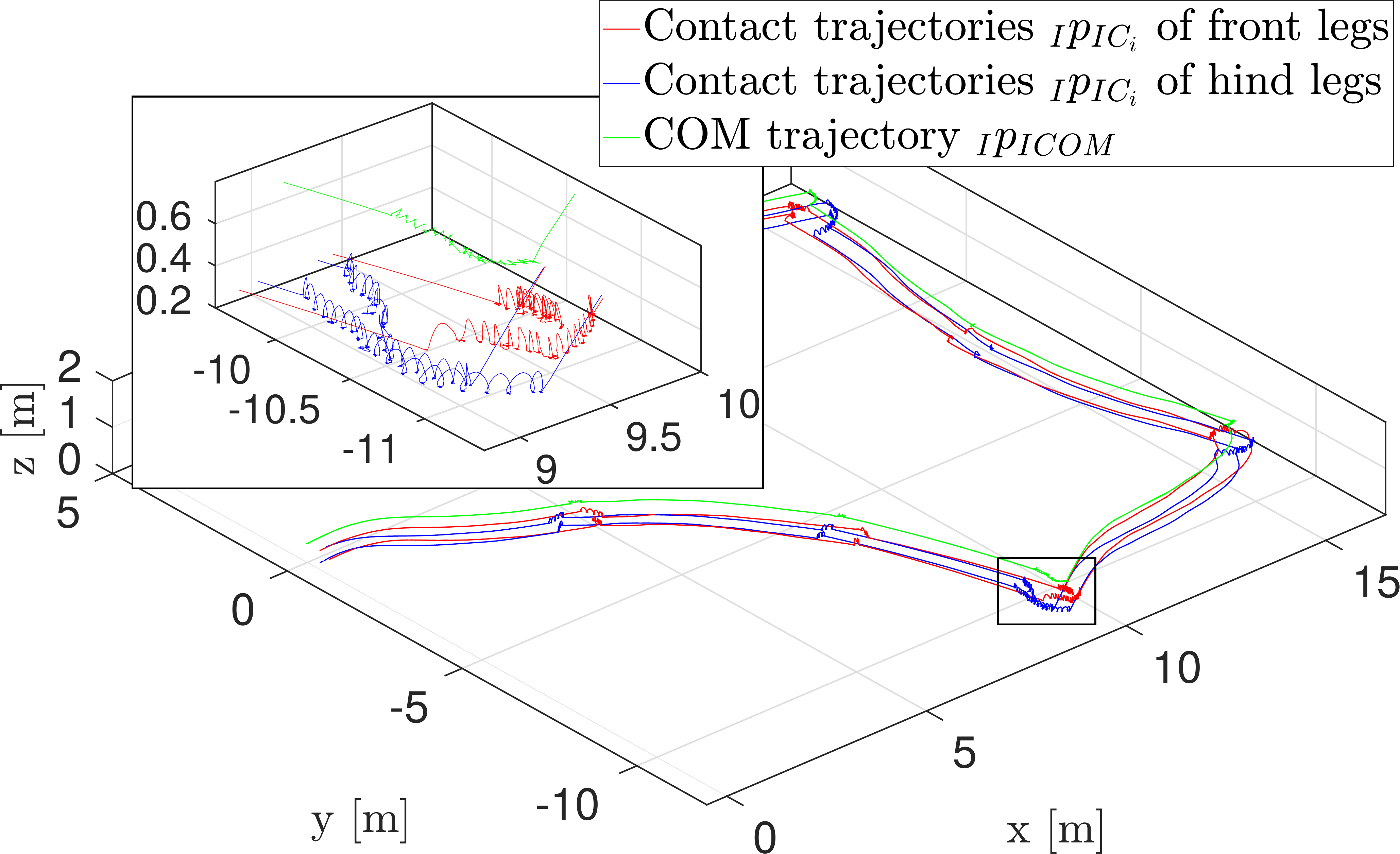}
    \caption{The robot ANYmal is driving and walking in an indoor environment (Available at \href{https://youtu.be/nGLUsyx9Vvc?t=103}{https://youtu.be/nGLUsyx9Vvc?t=103}). The three-dimensional plot shows estimated measurements of the robot where the red, blue and green lines depict the contact trajectories of the front legs, the contact trajectories of the hind legs, and the \ac{COM} trajectories \ac{w.r.t.} the inertial frame $I$. The zoomed-in figure shows transitions between driving and walking while the robot is performing a 90 degrees turn.}
    \label{fig:results_lab_trajectories}
    \vspace{-0.3cm}
\end{figure}
We performed driving and walking in an indoor environment, and the results are illustrated in Fig.~\ref{fig:results_lab_trajectories}. The three-dimensional plot shows the measured trajectories of the front legs, hind legs, and the \ac{COM}. In addition, the zoomed-in plot depicts the transitions between driving and walking in a corner. As discussed in~\cite{bjelonic2018skating}, the robot is able to drive small curvatures although the robot is equipped with non-steerable wheels. By yawing the base of the robot, the wheels are turning \ac{w.r.t.} an inertial frame. For larger curvatures, the robot needs to step. The results successfully prove the omnidirectional capabilities of the robot.
\subsection{Indoor Environment: Inclined Terrain}
\begin{figure}[t!]
    \centering
    \includegraphics[width=\columnwidth]{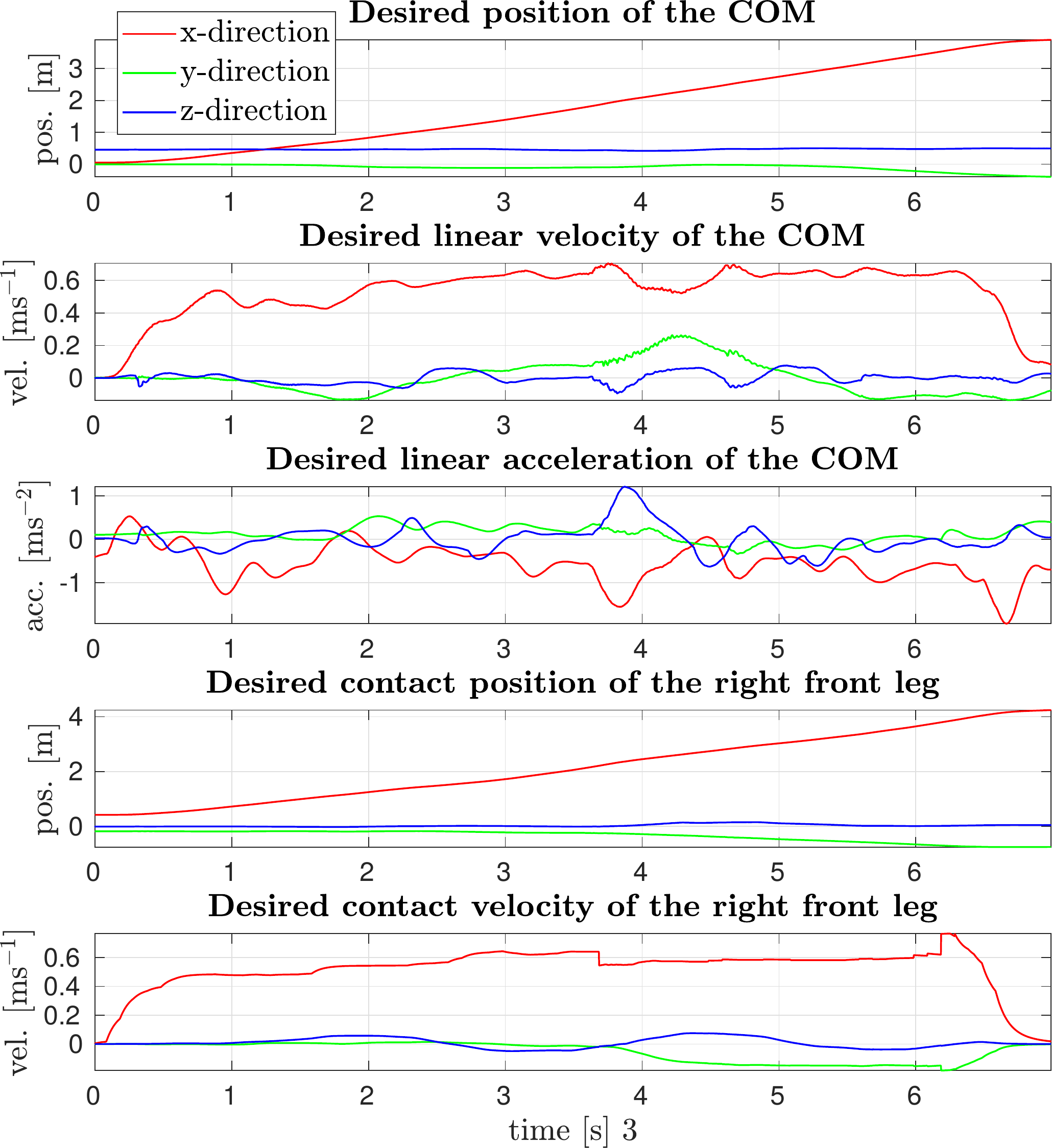}
    \caption{The plots show the desired motion (i.e., the optimized trajectories of the motion planner) of the \ac{COM} and the right front leg during the driving maneuver in Fig.~\ref{fig:double_incline} (Available at \href{https://youtu.be/nGLUsyx9Vvc?t=20}{https://youtu.be/nGLUsyx9Vvc?t=20}). The executed trajectories are almost identical to the desired motion shown here, and thus, the tracking error is negligible. This is due to the fast update rate (up to 200\,Hz) of the motion optimizer and the reinitialization of the optimization problem after every iteration with the measured state of the robot.}
    \label{fig:double_incline_opt_variables}
    \vspace{-0.5cm}
\end{figure}
Fig.~\ref{fig:double_incline} depicts the \ac{COM} motion tracked by the controller while ANYmal is driving blindly over two inclines and Fig.~\ref{fig:double_incline_opt_variables} illustrates the optimized trajectories of the motion planner while driving over the inclined terrain. Thanks to torque control, the robot adapts naturally to the unseen terrain irregularities while maintaining the \ac{COM} height. Moreover, the \ac{COM} motion is unaffected by the two obstacles although the robot drives at a speed of 0.7\,m/s. In addition, none of the wheels violates the friction constraints related to the no-slip condition.
\subsection{Outdoor Environment: Crossing a Street}
\begin{figure}[t]
    \centering
    \includegraphics[width=\columnwidth]{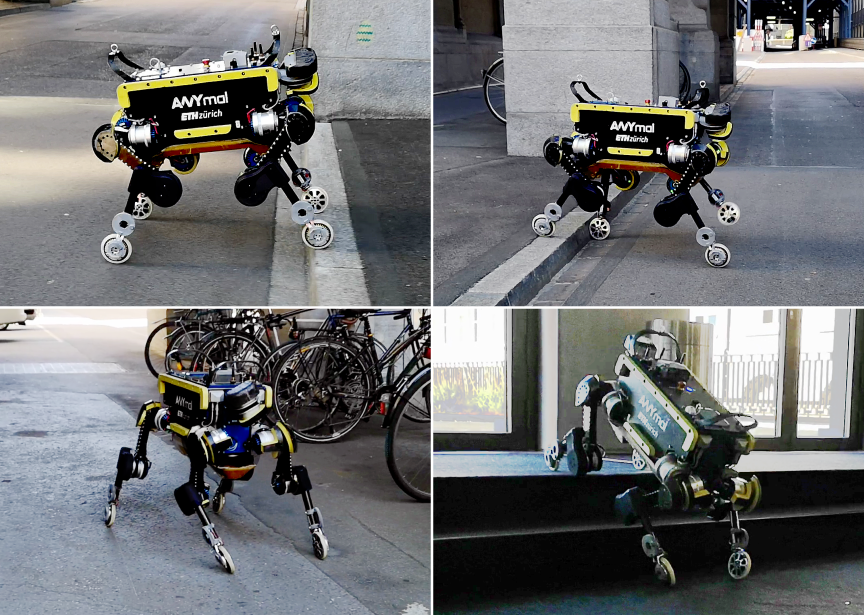}
    \caption{The figure illustrates several skills of the wheeled version of ANYmal (Available at \href{https://youtu.be/nGLUsyx9Vvc?t=38}{https://youtu.be/nGLUsyx9Vvc?t=38} and \href{https://youtu.be/nGLUsyx9Vvc?t=5}{https://youtu.be/nGLUsyx9Vvc?t=5}): dynamically driving down a step with 1\,m/s (top left image), walking up a step (top right image), driving in a curve by yawing the base (lower left image), and dynamically driving down stairs with 1\,m/s (lower right image).}
    \label{fig:result_figure}
    \vspace{-0.5cm}
\end{figure}

We conducted an outdoor experiment where we validated the performance of the robot under real-world conditions. Since the robot is able to drive fast and efficiently while being able to overcome obstacles, it applies to real-world tasks such as payload delivery. For this purpose, we conducted an experiment where the robot's task is to cross a street. As can be seen in Fig.~\ref{fig:result_figure}, the robot is able to drive down a step and to walk over another one. In addition, the lower left image illustrates how the robot rotates its base around the yaw direction to change its driving direction. This experiment also highlights the significant advantages of wheeled-legged robots compared to traditional walking robots. The robot is able to drive down steps with 1\,m/s without the need for terrain perception. Moreover, the lower right image of Fig.~\ref{fig:result_figure}, which shows the robot driving down a stair with 1\,m/s without the need to step, confirms the advantage.
\subsection{High Speed and Low Cost of Transport}
The computation of the mechanical \ac{COT} is based on the work in~\cite{bjelonic2018skating}. On flat terrain, the robot achieves a \ac{COT} of 0.1 while driving 2\,m/s and the mechanical power consumption is 63.64\,W. A comparison to traditional walking and skating with passive wheels~\cite{bjelonic2018skating} shows that the \ac{COT} is lower by 83\,\% \ac{w.r.t.} the trotting gait and by 17\,\% \ac{w.r.t.} skating motions. In addition, with 4\,m/s we broke ANYmal's maximum speed record of 1.5\,m/s given in~\cite{hwangbo2018learning}.
\section{CONCLUSIONS}
In this work, we show a whole-body motion control and planning framework for a quadrupedal robot equipped with non-steerable, torque-controlled wheels as end-effectors. The mobile platform combines the advantages of legged and wheeled robots. In contrast to other wheeled-legged robots, we show for the first time dynamic motions over flat and inclined terrains as well as over steps. These are enabled thanks to the tight integration of the wheels into the motion planning and control framework. For the motion optimization, we rely on a 3D ZMP approach which updates the motion plan continuously. This motion plan is tracked by a hierarchical \ac{WBC} which considers the nonholonomic contact constraint introduced by the wheels. Thanks to torque control, the robot does not violate the contact constraints and the fast update rates of the motion control and planning framework make the robot robust in the face of unpredictable terrain.

We aim to demonstrate further the application of the system to real-world tasks by conducting additional outdoor experiments. Future research will focus on hybrid locomotion strategies, i.e., walking and driving at the same time. To this end, promising initial results of a novel trajectory optimization for wheeled-legged quadrupedal robots further expand on the current motion planner presented by optimizing both \ac{COM} and foot trajectories in a single optimization using linearized \ac{ZMP} constraints~\cite{viragh2019trajectory}. In addition, perceptive motion planning over a long time horizon in challenging environments is still an unsolved problem for wheeled-legged and legged robots.






\section*{ACKNOWLEDGMENT}

The authors would like to thank Vassilios Tsounis for his support during the development of the wheel actuator firmware. Our gratitude goes to Francisco Gir\'aldez G\'amez and Christian Gehring who helped with the preliminary investigation of the rolling constraint. Furthermore, we thank Anna Beauregard for her comments on the final version of the paper.


\balance
\bibliographystyle{IEEEtran}
\bibliography{IEEEabrv,submissionbibfile}

\end{document}